\documentclass[sn-mathphys,Numbered]{sn-jnl}% Math and Physical Sciences Reference Style
\usepackage{graphicx}%
\usepackage{multirow}%
\usepackage{amsmath,amssymb,amsfonts}%
\usepackage{amsthm}%
\usepackage{mathrsfs}%
\usepackage[title]{appendix}%
\usepackage{xcolor}%
\usepackage{textcomp}%
\usepackage{manyfoot}%
\usepackage{booktabs}%
\usepackage{algorithm}%
\usepackage{algorithmicx}%
\usepackage{algpseudocode}%
\usepackage{listings}%
\usepackage{subfig}%
\usepackage{diagbox}%
\usepackage{orcidlink}%
\usepackage{natbib}%

\theoremstyle{thmstyleone}%
\theoremstyle{thmstyletwo}%

\theoremstyle{thmstylethree}%

\usepackage{hyperref}
\hypersetup{
    colorlinks=true,
    linkcolor=blue,
    citecolor=blue,
    urlcolor=blue
}

\begin{document}

\title[Article Title]{Learner Attentiveness and Engagement Analysis in Online Education Using Computer Vision
}


\author*[1]{\fnm{Sharva} \sur{Gogawale},\orcidlink{0009-0000-5230-5197}}\email{sharvag@mail.tau.ac.il}
\equalcont{These authors contributed equally to this work.}

\author[2]{\fnm{Madhura} \sur{Deshpande}}\email{madhurad@mail.tau.ac.il}
\equalcont{These authors contributed equally to this work.}

\author[3]{\fnm{Parteek} \sur{Kumar}}\email{parteek.bhatia@thapar.edu}

\author[2]{\fnm{Irad} \sur{Ben-Gal}}\email{bengal@tauex.tau.ac.il}

\affil*[1]{\orgdiv{School of Electrical Engineering}, \orgname{Tel-Aviv University}, \orgaddress{\city{Tel Aviv}, \postcode{6997801}, \country{Israel}}}

\affil[2]{\orgdiv{Department Industrial Engineering}, \orgname{Tel-Aviv University}, \orgaddress{\city{Tel Aviv}, \postcode{6997801}, \country{Israel}}}

\affil[3]{\orgdiv{Department of Computer Science and Engineering}, \orgname{Thapar Institute of Engineering and Technology}, \orgaddress{\city{Patiala}, \postcode{147004},\country{India}}}

\abstract{In recent times, online education and the usage of video-conferencing platforms have experienced massive growth. Due to the limited scope of a virtual classroom, it may become difficult for instructors to analyze learners' attention and comprehension in real time while teaching. In the digital mode of education, it would be beneficial for instructors to have an automated feedback mechanism to be informed regarding learners' attentiveness at any given time. This research presents a novel computer vision-based approach to analyze and quantify learners’ attentiveness, engagement, and other affective states within online learning scenarios. This work presents the development of a multiclass multioutput classification method using convolutional neural networks on a publicly available dataset - DAiSEE. A machine learning-based algorithm is developed on top of the classification model that outputs a comprehensive attentiveness index of the learners. Furthermore, an end-to-end pipeline is proposed through which learners' live video feed is processed, providing detailed attentiveness analytics of the learners to the instructors. By comparing the experimental outcomes of the proposed method against those of previous methods, it is demonstrated that the proposed method exhibits better attentiveness detection than state-of-the-art methods. The proposed system is a comprehensive, practical, and real-time solution that is deployable and easy to use. The experimental results also demonstrate the system’s efficiency in gauging learners’ attentiveness.
}

\maketitle

\section{Introduction}\label{sec1}

The Covid-19 pandemic has significantly transformed the learning approaches of both instructors and learners, as their educational settings have shifted from traditional physical classrooms to digital ones. Millions of people globally use platforms such as Google Meet, Zoom, Teams, and WebEx to deliver lessons, conduct one-on-one sessions, and arrange seminars or conferences. E-learning has become substantially popular due to its feasibility, high demand, and the multitude of benefits that it offers. Although e-learning provides learners with flexible and self-paced learning schedules, it also results in certain inevitable challenges \cite{1}. 

In the traditional offline model of education, instructors can analyze learners’ engagement and responses to the material being taught by observing their body language, head movement, and various facial cues. Past studies have proposed that learning comprises a diverse range of cognitive and affective states. These studies have recognized facial expressions that are linked to learning-centered states, which typically include boredom, confusion, engagement, and frustration \cite{2}. In the online mode of education, there is an absence of a real-time feedback mechanism between the learners and instructors. This model lacks a real-time feedback mechanism between the learners and instructors, making it extremely difficult for instructors to analyze learners’ engagement levels and attentiveness. As a result, instructors are often unable to adjust their teaching based on learners’ visual responses.Consequently, the learning experiences and outcomes could be negatively impacted. 

Understanding the affective states of learners can facilitate more effective guidance and allow instructional design to be tailored to their needs. This can also help learners who require assistance and could reduce dropout rates in online learning \cite{3}. Enhancing learner performance and engagement is crucial for online learning. Therefore, there is a pressing need to develop more affordable and accessible methods for capturing, monitoring and analyzing learners’ attention. By doing so, the effectiveness of these environments will be maximized and they will be conducive for learners to achieve their educational goals \cite{5}.
Engagement detection prevails in multiple domains and it is done using various surveys, eye trackers, head movements, gaze detection, and gauging the seven universal emotions viz. happiness, surprise, contempt, sadness, fear, disgust, and anger. In the education domain, it is crucial to understand how learners acquire, comprehend, and analyze information. The brain’s function is contingent upon the integration of both cognitive and affective processing. The affective state of a learner has proved to play a very significant role. Holistic improvement in creative thinking is observed in individuals experiencing positive emotional states, whereas those in negative emotional states prioritize setting goals for performance (well-being) over mastery (learning) \cite{6}.

Facial image analysis is the most highly recognized external factor for detecting facial expressions and affective states. Facial features can be used to identify the key emotional states that are represented by learners’ expressions during learning \cite{7}. Although this area has exhibited some advancements, multiple challenges exist, particularly in accurately associating facial expressions with the underlying emotions of learners. Advances in neural networks and deep learning techniques have enabled the visual perception of machines to significantly progress, but only a few efforts have been made to detect learners’ attentiveness in online learning.

In this research, a hybrid architecture using a convolutional neural network (CNN) is presented. This is a popular deep-learning approach that has been used to analyze learners’ aggregated affective state patterns, including understanding learning-centered emotions such as boredom, engagement, confusion, and frustration, and estimating their attentiveness states throughout the online session. Furthermore, a novel comprehensive framework is proposed for implementing a low-cost system for emotion recognition and attentiveness estimation, which can be integrated into video conferencing platforms or e-learning platforms. Since instructors require external perceptible factors to gauge perceived attentiveness, this automation can provide critical insight into their learners’ learning experiences.

The remainder of this paper is organized as follows. A brief overview of engagement estimation is provided in Section \ref{sec2}. The methods implemented in this research, as well as descriptions of the used dataset and models, are introduced and discussed in Section \ref{sec3}. The development of the Attentiveness Index is outlined in Section \ref{sec4}. The end-to-end pipeline is illustrated in Section \ref{sec5}. The performance evaluation results of our proposed method compared with other recent works are discussed in Section \ref{sec6}. Finally, the conclusions and potential future extensions of the proposed framework are presented in Section \ref{sec7}.

\section{Related Work}\label{sec2}

Over the last few years, extensive efforts have been dedicated to detecting and analyzing learner engagement \cite{4,53}. This ongoing research on automated cognitive engagement detection in online classrooms is crucial for adapting to the unique challenges and opportunities presented by the digital mode of education \cite{8}. At a general level, automated methods for detecting learners’ engagement can generally be categorized into computer vision-based methods and multimodal methods. Multimodal techniques are typically employed for identification tasks and utilize expensive equipment for engagement detection.

 Yang et al. \cite{9} used an eye-tracking system (FaceLAB 4.5) to observe and evaluate the variations and patterns of visual attention in university learners during PowerPoint presentations. Similarly, Kao et al. \cite{10} used the ASL Mobile Eye-XG eye tracker to investigate learners' engagement while reading. The eye-tracking data was wirelessly transmitted to a system for analysis. Leite et al. \cite{43} implemented both manual and automation techniques to gauge disengagement by analyzing different types of data, namely audio, visual, contextual, etc. Kapoor et al. \cite{44} extracted multimodal sensory information through postural shifts and facial expressions measured using a camera and pressure-sensing chair to classify children's affect. 

Another studied approach for estimating engagement involves integrating and analyzing biological data gathered from a range of sensors, including EEG, heart rate, blood pressure, and galvanic skin response. Fairclough et al. \cite{45} analyzed data collected through EEG, ECG, respiratory rate, skin conductance level etc. to analyze it with respect to three subjective states, namely task engagement, distress and worry. Monkaresi et al. \cite{47} examined heart rate measurements and Chaoachi et al. \cite{46}  utilized EEG in a learning environment to examine the relationship between learners’ engagement and emotional states.

Most methods for capturing and monitoring learners' engagement rely significantly on expensive, commercial, or specialized equipment. These approaches are typically conducted offline, utilizing data collected from dedicated devices installed in classrooms or labs.

Identifying the precise factors for accurately measuring attentiveness has also become a critical issue. Visual information methods contain a bigger scope to study multiple features, which are also easily accessible. Gaze estimation methods are the most useful visual information methods. Ishikawa et al. \cite{11} employed a generative face model to track the complete head. This model was used to robustly extract eye corners, the eye regions, and the head poses to build a  monocular driver gaze tracking system. Wood et al. \cite{13}  introduced EyeTab, an approach that adapts conventional computer vision algorithms for near-real-time binocular gaze estimation compatible with unmodified portable tablets. It performed multistage eye localization to identify eye centers and refined eye regions of interest (ROIs) and further demonstrates ellipse fitting methods for gaze estimation. Wood et al. \cite{12}  proposed the first morphable eye model that used eyeball movement to achieve high gaze estimation accuracy in challenging user-contingent settings. Valenti et al. \cite{14} focused on a hybrid approach that combined head pose and eye location information for computing gaze direction. However, these gaze estimation methods rely on specialized equipment such as high-quality cameras, infrared lighting, and sensors that measure the distance, however it incurs high economic and spatial costs.

Computer vision-based methods are mainly categorized into two approaches, namely feature-based and end-to-end. Feature-based approaches involve extracting specific features from images or videos, and then processing and analyzing these features to measure the engagement level by using a classifier or regressor \cite{15, 16, 17, 18}. 

Krithika et al. \cite{27} processed eye and head movement information to study concentration levels and provided low-concentration alerts. Grafsgard et al. \cite{19})  used the Computer Expression Recognition Toolbox (CERT) to track facial movements within a natural video compilation. Whitehill et al. \cite{20}  proposed different feature extraction techniques and classifier combinations to detect the level of engagement of learners from image data. Wu et al. \cite{17}  employed a function for engineering numerous aligned features for learner engagement intensity prediction by utilizing the EmotiW2020 dataset. They combined  long short-term memory (LSTM), a gated recurrent unit (GRU), and a fully connected layer to process motion features from videos captured under several varying conditions. Zhu et al. \cite{18} used attention-based hybrid deep models to predict the engagement intensity of students watching educational videos. They proposed a robust ensemble model incorporating a GRU network with multi-rate and multi-instance attention.

Furthermore, researchers have used various models to analyze extracted features. Zhaletelj et al. \cite{21}  and Whitehill et al. \cite{22}  used a machine learning model to analyze extracted features, and Yang et al. \cite{23} and Niu et al. \cite{24} used LSTM and GRU to analyze the time series in videos. Monkaresi et al. \cite{25} used a combination of various features such as heart rate, animation features (from Microsoft Kinect Face Tracker), and local binary patterns to detect engagement in students during and after performing a writing activity.

End-to-end models utilize a deep CNN classifier to detect engagement by feeding the raw frames of videos or images. Gupta et al. \cite{29} introduced DAiSEE which is a multilabel video classification dataset that can be used for identifying learners’ affective states. They employed existing network architectures and convolutional video classification techniques such as InceptionNet \cite{30}, C3D  \cite{31} , and LRCN \cite{32}. Geng et al. \cite{33} implemented the C3D network and focal loss function to address the class imbalance issue in the dataset. Zhang et al. \cite{34} modified the existing Inflated 3D Convolutional Network (I3D) \cite{26}  network to use 3D convolution for handling both spatial and temporal features of learners’ videos to analyze the complete learning process. Liao et al. \cite{35} introduced the deep facial spatio-temporal Network (DFSTN) and evaluated it for learners’ engagement detection in online learning on the DAiSEE dataset. Dewan et al. \cite{36}  used a face recognition descriptor to calculate each learner’s edge features and further implement kernel principal component analysis to allow separability and examine the correlation between the extracted features. Their work focused on the ‘engagement’ label of the DAiSEE dataset and provided two-level and three-level engagement decisions. Similarly, Murshed et al. \cite{37} modified the four-level engagement affective state in DAiSEE to provide a three-level decision for engagement analysis. Their method involved implementing a CNN model by assimilating useful features of three existing networks.

Sharma et al. \cite{28} proposed a method using eye movement and facial expression detection to analyze concentration levels in e-learning environments. The learners’ emotions were categorized into seven universal emotion, happiness, surprise, contempt, sadness, fear, disgust, and anger. However, Lehman et al. \cite{38} confirmed that basic emotions were rare in learning sessions, which is particularly challenging for methods relying on facial expressions. In educational settings, gauging and detecting learning-centered emotion in real-time is important.

 Few prior studies have focused on learning- centered emotions in real environments, instead focusing more focusing on analyzing images for engagement rather than video analysis in practical settings. A cost-effective, accessible, and easy-to-use implementation of the engagement detection pipeline is also lacking.  The present study aims to address these gaps using computer vision techniques and deep learning networks to build an automated feedback mechanism for informing instructors about learners’ comprehensive attentiveness in real time. This study also proposes the implementation of an end-to-end online system that illustrates dynamic interfaces for learners as well as instructors through which a comprehensive attentiveness analysis can be conducted in real time in online learning environments.

\begin{figure*}
    \centering
    \includegraphics[width=\linewidth]{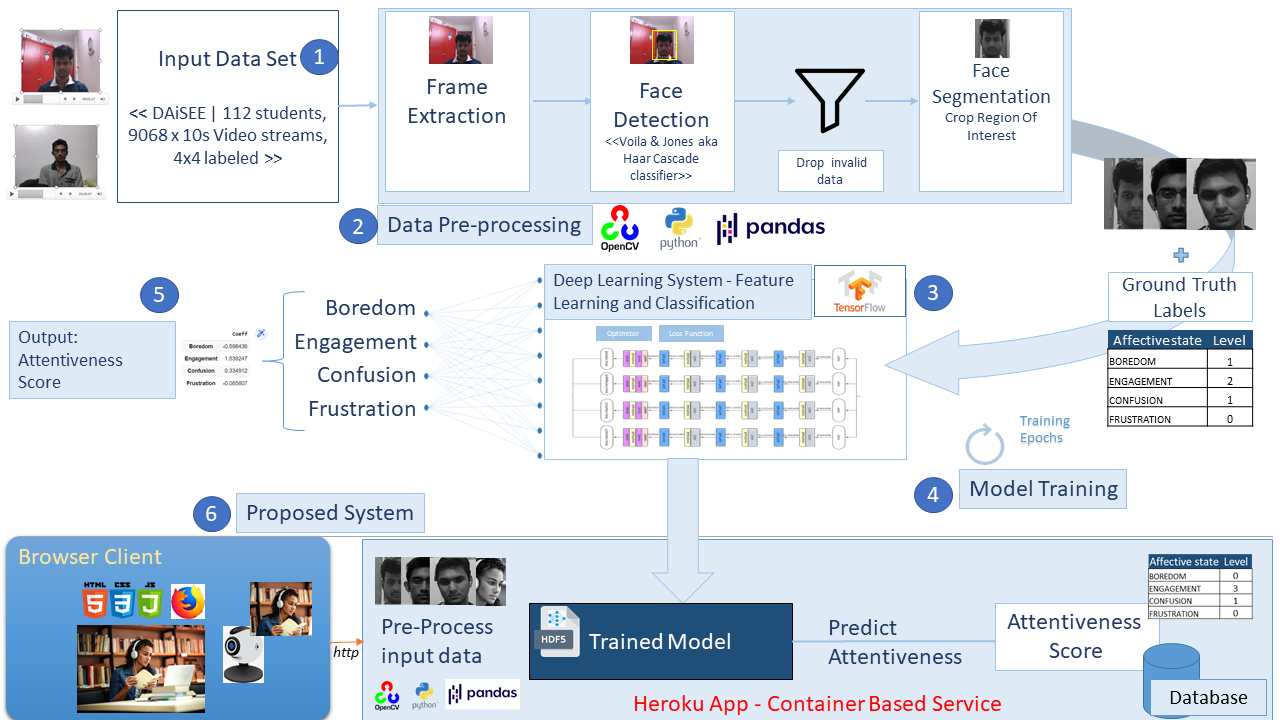}
    \caption{Block diagram illustrating the overall workflow of the proposed pipeline. Stages include data selection, data preprocessing, model training, attentiveness index calculation, and web client interface deployment}
    \label{fig:pl}
\end{figure*}

\section{Implementation of Proposed System}\label{sec3}

Fig: \ref{fig:pl} illustrates the implementation of the proposed system. The chosen publicly available dataset is suitable for classifying affective states. This dataset undergoes a preprocessing phase to make it suitable for feeding into the network architecture. The model is trained on these processed data, generating predictive outcomes. An attentiveness formula evaluates and quantifies the attentiveness of learners by employing classic machine learning techniques. This mathematical formula connects the raw model outputs to a comprehensive attentiveness index. The proposed framework of the system uses live feeds of learners as its primary input source and provides comprehensive attention and engagement analytics to the instructor in real time. The subsequent sections detail each part of the implementation is discussed in detail in the subsequent sections, providing a comprehensive view of the system’s development and functionality.

\subsection{Dataset}\label{subsec1}
\begin{figure}
    \centering
    \includegraphics[width=\linewidth]{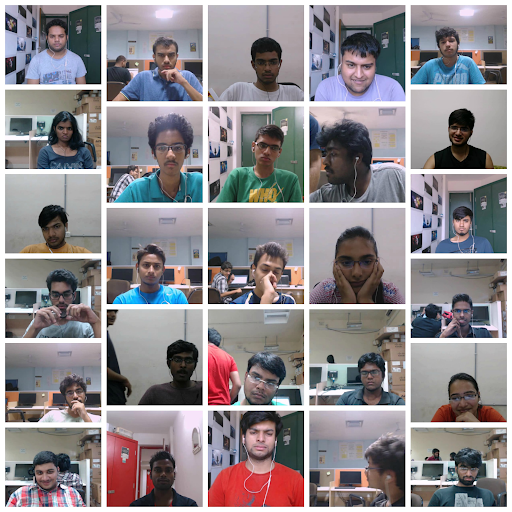}
    \caption{Snapshot of DAiSEE Dataset}
    \label{fig:daisee_ds}
\end{figure}

The proposed attention estimation approach is evaluated using the open-source dataset, DAiSEE \cite{39}. DAiSEE is the first multilabel video classification dataset to include 9068 video snippets from 112 individuals (32 females and 80 males) captured in “in the wild” scenarios for recognizing learners’ states, such as boredom, confusion, engagement, and frustration. These videos were captured at a resolution of 1920 × 1080 pixels at 30 frames per second (fps), through a full HD webcam connected to a computer. Each video is 10 seconds long and composed of 300 frames. The dataset includes four levels of intensity labels, very low (0), low (1), high (2), and very high (3), for each of the affective states, i.e., boredom, engagement, confusion, and frustration. These labels are crowd-annotated and correlated with a gold standard annotation created using a team of expert psychologists. The broad spectrum of possible combinations of these labels helps to describe the overall cognitive state of the learner. Fig: \ref{fig:daisee_ds} shows a snapshot of the dataset.

Table \ref{table:1} highlights the exact number of videos and the label categorization in the training (total videos: 5358), testing (total videos: 1784), and validation (total videos: 1429) sets of DAiSEE.

\begin{table}
\caption{DAiSEE Dataset Distribution} 
\label{tab:title} 
\centering

\begin{tabular}{||c| c c c c||} 
 \hline
 Boredom & 0 & 1 & 2 & 3 \\ [0.5ex] 
 \hline\hline
 Training & 2433 & 1696 & 1073 & 156\\ 
 \hline
 Validation & 446 & 376 & 475 & 132\\
 \hline
 Testing & 823 & 584 & 338 & 39\\
 \hline
\end{tabular}

\hspace{1em} % Adjust the space as needed

\begin{tabular}{||c| c c c c||} 
 \hline
 Engagement & 0 & 1 & 2 & 3 \\ [0.5ex] 
 \hline\hline
 Training & 34 & 213 & 2617 & 2494\\ 
 \hline
 Validation & 23 & 143 & 813 & 450\\
 \hline
 Testing & 4 & 84 & 882 & 814\\
 \hline
\end{tabular}

\hspace{1em} % Adjust the space as needed

\begin{tabular}{||c| c c c c||} 
 \hline
 Confusion & 0 & 1 & 2 & 3 \\ [0.5ex] 
 \hline\hline
 Training & 3616 & 1245 & 431 & 66\\ 
 \hline
 Validation & 942 & 322 & 153 & 12\\
 \hline
 Testing & 1200 & 427 & 136 & 21\\
 \hline
\end{tabular}

\hspace{1em} % Adjust the space as needed

\begin{tabular}{||c| c c c c||} 
 \hline
 Frustration & 0 & 1 & 2 & 3 \\ [0.5ex] 
 \hline\hline
 Training & 4183 & 941 & 191 & 43\\ 
 \hline
 Validation & 1058 & 271 & 81 & 19\\
 \hline
 Testing & 1388 & 316 & 57 & 23\\
 \hline
\end{tabular}

\label{table:1}
\end{table}

\subsubsection{Data Imbalance Issue}\label{subsubsec1}

An important issue observed in Table \ref{table:1} is the imbalanced distribution of data across different affective states in DAiSEE. The support for the ‘very low’ and ‘low’ engagement labels is significantly lower than the support for the ‘very high’ engagement labels. However, for boredom, confusion, and frustration, the opposite is true, i.e., the support for lower-intensity labels is much higher than that for the higher-intensity labels.
Conventional machine learning techniques are typically modeled to optimize the total classification accuracy \cite{15}. Their performance is typically optimal when given a balanced data distribution. As DAiSEE is an imbalanced dataset, the classifier tends to be biased toward the majority class, reducing the accuracy for the minority class, and making it difficult to extract valuable information and features related to the minority classes. Consequently, the imbalanced dataset is prone to overfitting. Two types of solutions that can be implemented to address such class imbalance issues are data-based solutions and algorithm-based solutions. In this research, focal loss is employed to reduce the bias issue that comes with class-imbalanced data distribution.

\subsection{Data Preprocessing}\label{subsec2}

\begin{figure}
    \centering
    \includegraphics[width=\linewidth]{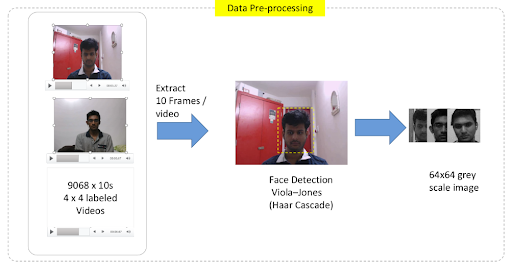}
    \caption{Data Preprocessing of video files}
    \label{fig:dpp}
\end{figure}

Data pre-processing refers to the processing of the input video for feeding into the classification model. The 30 fps video snippets are taken to extract the image frames. Further, the Viola-Jones Face Detection (popularly known as Haar Cascade) is employed to extract the region of interest from the image frame. OpenCV's Haar cascade frontal face module is applied to the input frames to achieve this task. The facial landmarks and pupils were detected correctly for most of the images. The images are marked invalid and discarded from the network, where the facial landmarks and features cannot be extracted. The detected face areas are resized to a uniform size, and the RGB channels are combined into normalized grayscale images of size 64x64. This reduces the dimensionality and computational complexity of the data.
The pipeline further processes these inputs.Fig: \ref{fig:dpp} illustrates the data pre-processing method.

\subsection{Model Architecture}\label{subsec3}
A novel approach involving an \emph{n} parallel (here \emph{n}=4) branched model has been used for building the classifier. A separate classifier has been built for each affective state corresponding to each branch. Each classifier predicts the intensity level of that particular affective state. Combining predictions from all these classifiers, the overall output label for the input frame is obtained that contains the corresponding intensities of each affective state. The following sections illustrate the models built using 4 parallel branches.

\subsubsection{CNN based Architecture}\label{subsubsec1}

\begin{figure}
    \centering
    \includegraphics[width=\linewidth]{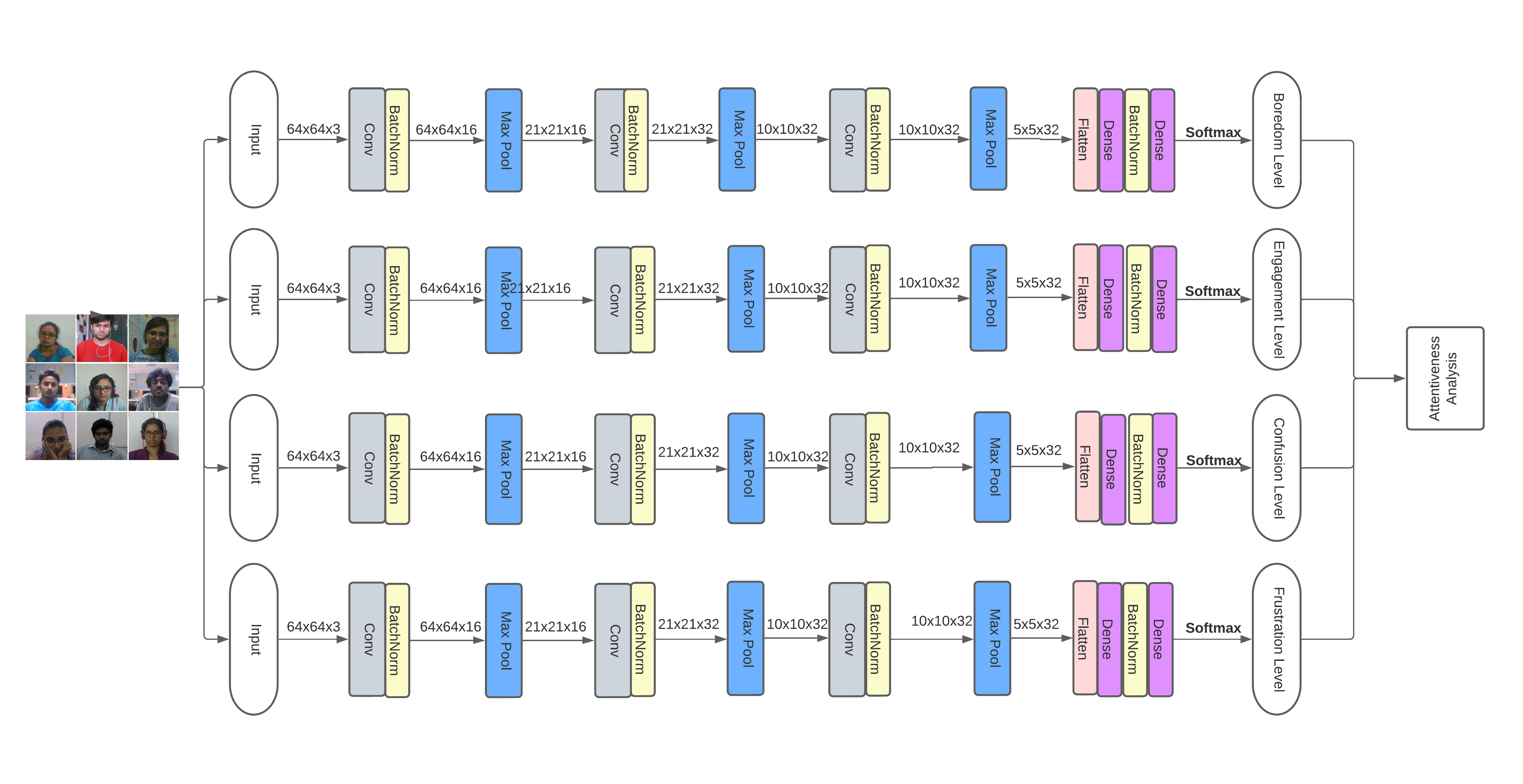}
    \caption{The architecture of the hybrid CNN network}
    \label{fig:CNN}
\end{figure}

In this work, a 4 parallel-branched CNN-based architecture is used as a feature extraction network for
classifying the affective states. It is a lightweight model having 0.47 M parameters. The 4 branches are used for the 4-level classification. The final softmax layer of each branch provides a single vector of size 1 X 4 representing the intensity values for each level.
The $i^{\text{th}}$ element in the vector output of each branch represents the predicted probability that the learner in the input image is at the $i^{\text{th}}$ level exhibiting that specific affective state where 0 indicates disengagement and 3 indicates very high engagement.
The network's final output is the concatenation of the outputs from all 4 branches, a vector of size 4 X 4 containing the scores of each level from class, i.e., boredom, confusion, engagement, and frustration.
Fig: \ref{fig:CNN} illustrates the model architecture using CNN.

\subsubsection{EfficientNet based Architecture}\label{subsubsec2}

\begin{figure}
    \centering
    \includegraphics[width=\linewidth]{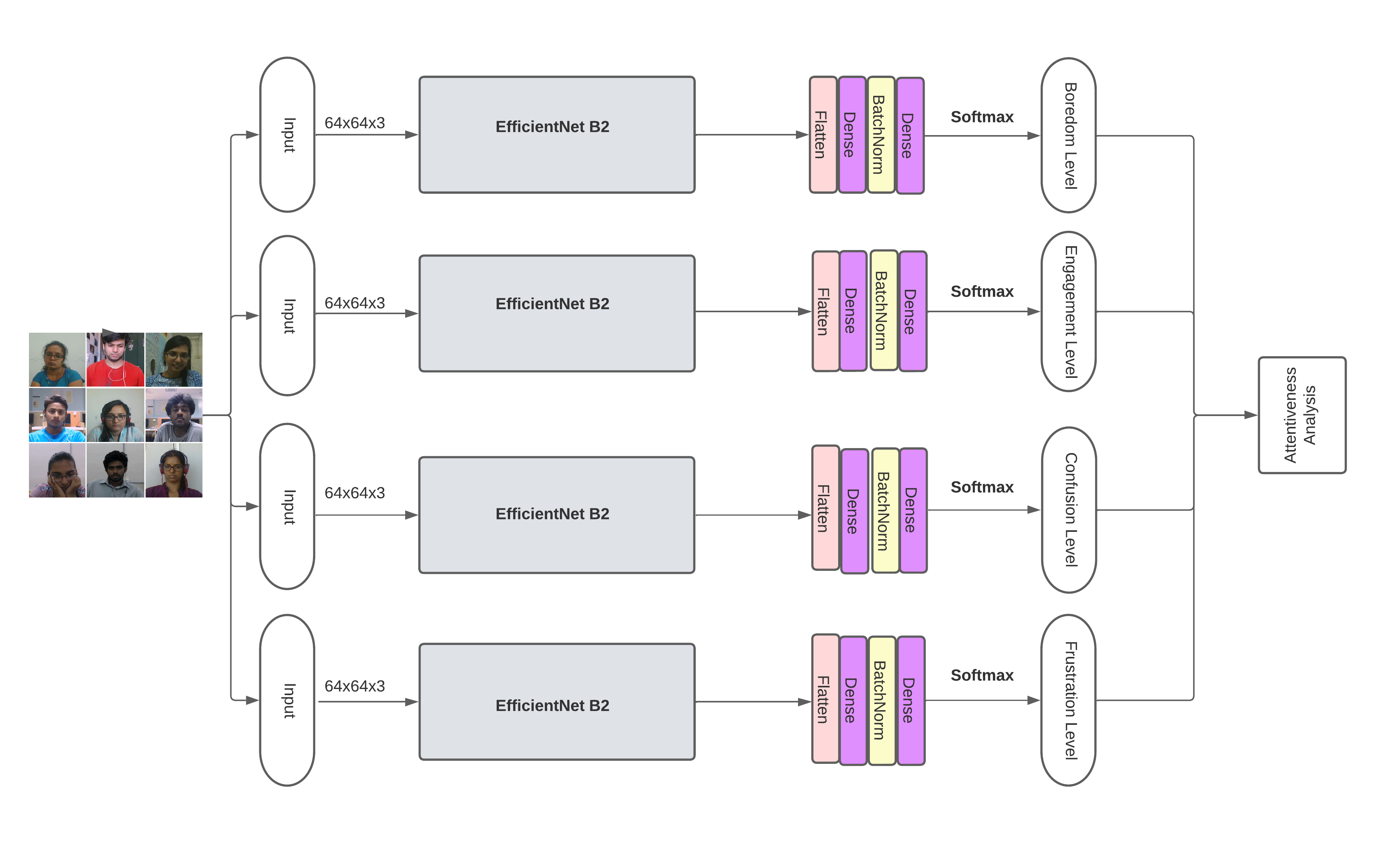}
    \caption{The architecture of the hybrid EfficientNet network}
    \label{fig:EfficientNet.png}
\end{figure}

A similar 4 parallel branch EfficientNet (EfficientNetB2) model has been used as a backbone network for feature extraction for classifying the affective states. It has 31.7 M parameters making it a lightweight model easy for cloud deployment compared to other networks such as VGG, ResNet etc. The main building block of this architecture \cite{48} is mobile inverted bottleneck MBCon to which squeeze-and-excitation optimization is added. The final softmax layer of each branch provides a single vector of size 1 X 4 representing the intensity values for each affective state.
Fig: \ref{fig:EfficientNet.png} illustrates the model architecture using EfficientNet.

\subsection{Loss Function}\label{subsec4}
In this paper, categorical focal loss is employed to address the issue of class imbalance in the data.
Focal Loss \cite{49} applies a modulating term to the categorical cross-entropy loss to focus learning on hard negative examples and improves the classification performance. 
Focal Loss is represented as:
\begin{equation} \label{eq:focal_loss}
FocalLoss(p_t) = \alpha \cdot (1 - p_{t})^\gamma \cdot CategoricalCE(y_{true}, y_{pred})
\end{equation}

Here $p_t$ is given by, 
\begin{equation} \label{eq:pt_definition}
p_{t} = \begin{cases} output ,& \text{if } y_{true} == 1\\
 1 - output, & \text{otherwise}
\end{cases}
\end{equation}

and CategoricalCE is given by, 
\begin{equation} \label{eq:categorical_ce}
CategoricalCE(y_{true}, y_{pred}) = - \sum_{n=1}^{C} y_{{true}_i} * log(y_{{pred}_i})
\end{equation}

$\alpha$ is a balancing factor that can be adjusted to assign more or less weight to different classes and $\gamma$ is the focusing parameter that controls how much the loss focuses on difficult examples.

\subsection{Model Training}\label{subsec5}
The deep learning models are compiled using Keras (a high-level API) with TensorFlow 2.0 backend using the Python environment. Model was trained on a machine with Intel core CPU and Nvidia GTX GPU. Adam Optimizer with the initial learning rate set to 0.001 was used. The training was done with an early stopping criterion to obtain the model  weights from best performing epoch. Alpha balanced focal cross entropy loss has been used in this process for 4-level classification.

\section{Developing Attentiveness Formula}\label{sec4}

Classifying only affective states cannot comprehensively depict the learners’ attentiveness. Introducing a single parameter that accounts for these   affective states and instantly helps instructors understand whether the learner is attentive is crucial. For this purpose, a mathematical model has been developed that connects affective states using  an Attentiveness Index (Ai). The main tasks are to find corresponding appropriate weights for each affective state, to understand how the affective states impact the overall index (whether negatively or positively, and to what extent), and validate the model using the current cognitive science literature.
For this purpose, a subset of the  dataset   was   evaluated   and crowd-annotated by multiple instructors. Each video was given an attentiveness score on a scale of 1 to 10 (1 being the least attentive and 10 being the most attentive). The average attentiveness score was calculated based on all the labels provided by instructors' assessments. A machine learning algorithm based on multiple linear regression was applied on top of this newly formed   data subset provided the weights for each affective state, consequently providing the overall mathematical formula linking the affective states to a unique attentiveness index.

The obtained mathematical formula is :
\begin{equation} \label{eq:attentiveness_index}
A_i = -0.598 \cdot B + 1.539 \cdot E + 0.334 \cdot C - 0.085 \cdot F
\end{equation} where A\textsubscript{i} is the Attentiveness Index and B, E, C, and F represent the intensity values for Boredom, Engagement, Confusion, and Frustration respectively.
This formula \eqref{eq:attentiveness_index} illustrates the   weight   of each affective state and whether it contributes positively or negatively to the attentiveness index.
The resulting weights obtained for each of the affective states contributing to overall attentiveness align with past research propositions. Persistent observations across various learning environments indicate that boredom is linked to suboptimal learning outcomes, whereas frustration appears to have a weaker association with poor learning. Learning environments most often exhibit boredom and engaged concentration. The coefficient trends obtained for engagement and confusion are positive, with the confusion coefficient being less positive than the engagement coefficient. As per the literature, confusion positively impacts learning, and the positive coefficient, although not overly high, aligns with the literature. Moreover, boredom and frustration exhibit negative weights, where the boredom coefficient is more negative. These trends align with the existing cognitive studies \cite{40, 41, 42}.

\section{Online Platform for Attentiveness Analysis}\label{sec5}
Fig \ref{fig:pl} depicts the primary components of the proposed system for assessing learners’ affective states and engagement. Instructors can employ this system during live lecture sessions in virtual classrooms, remote learning settings, or any other learning environments that use personal computers and web cameras.
Two types of participants are present in the education setting: the instructor and the learner. When the learner interacts with learning material, the system automatically analyzes the live video feed of the learner to evaluate their attentiveness, engagement, and affective states. The system provides this analysis to the instructors through visual and tabular feedback. 

\begin{figure}
    \centering
    \subfloat[Graphical]{\includegraphics[width=\linewidth]{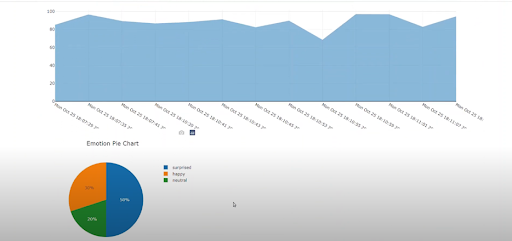}\label{fig:graph}}
    
    \subfloat[Tablular]{\includegraphics[width=\linewidth]{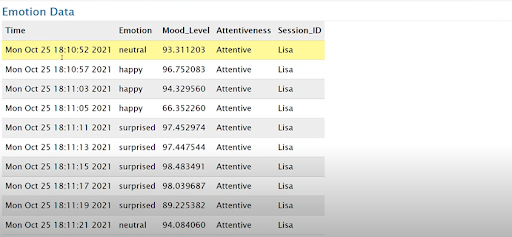}\label{fig:table}}
    
    \caption{Examples of illustrations of graphical and tabular analytics captured during the system's different prototype stages}
    \label{fig:examples}
\end{figure}

\begin{figure}
    \centering
    \includegraphics[width=\linewidth]{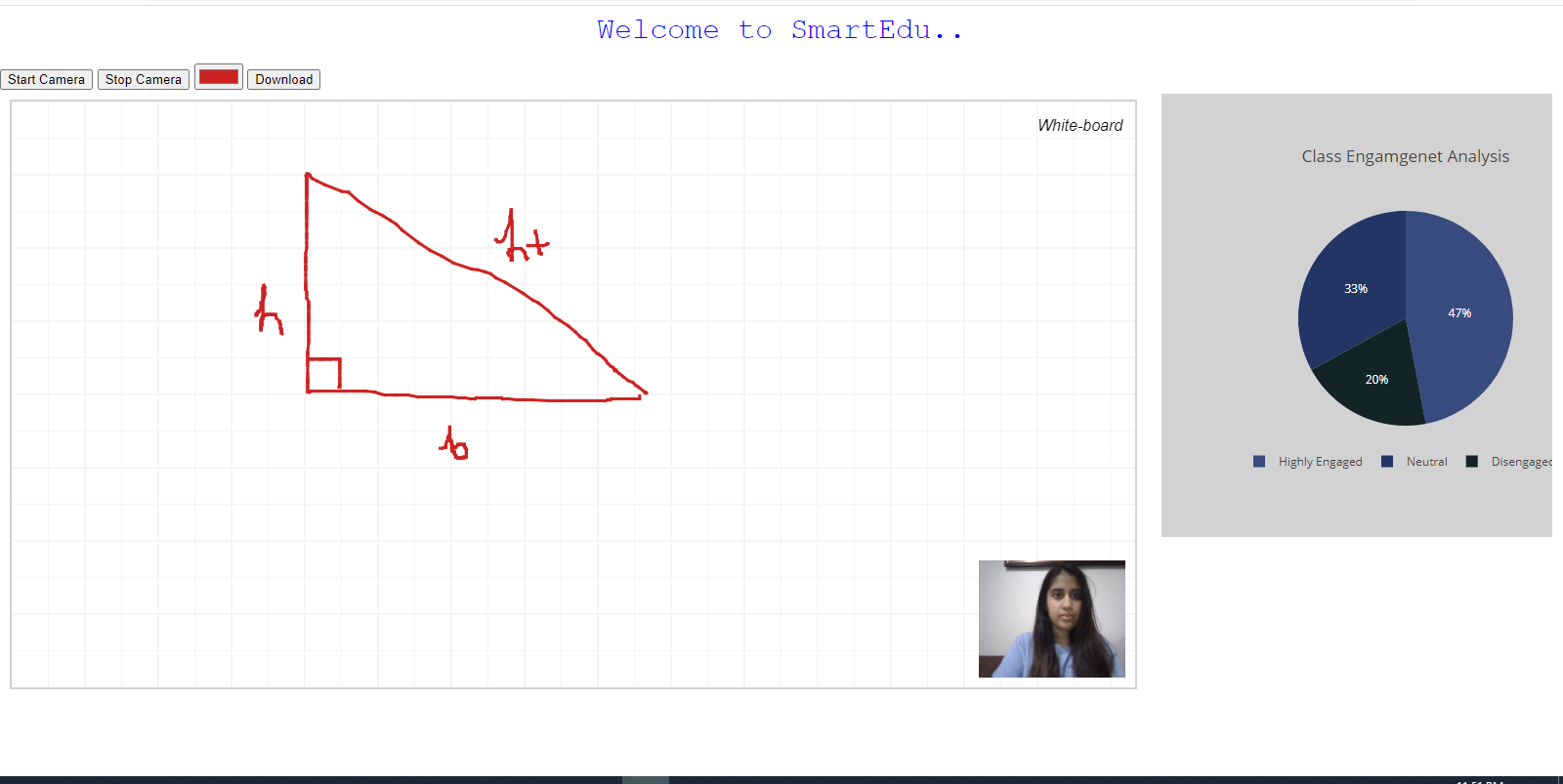}
    \caption{Prototype of Instructor's Interface}
    \label{fig:teacher_if}
\end{figure}

The main features of the proposed system are as follows:

\begin{itemize}
\item The system includes two types of dynamic interfaces-: one for instructors and one for learners.
\item The system is built and deployed on a cloud server.
\item The system supports multiple-user login simultaneously.
\item It can be accessed through this link: \url{https:// smart-edu-system-01.herokuapp.com}
\item Fig \ref{fig:examples} depicts examples of the system’s outputs, i.e., the engagement and attentiveness analytics.
\item The analytics can be seen are depicted on on the instructor’s system interface part of the system, as shown in Fig \ref{fig:teacher_if}
\item During online sessions, if the accumulated engagement of the class falls below a certain threshold, instructors  receive alerts (e.g., class is currently disengaged!)
\item Instructors are provided with detailed visual and graphical analytics in the lecture context.
\item Through the system, instructors can identify the exact point in the lecture where the attentiveness levels dropped throughout the lecture (e.g., attentiveness levels were low (30\%) at the end of the lecture).
\item They can also view whether certain affective states were high during any part of the lecture. (e.g., confusion levels were high (70\%) while the instructor was explaining a difficult concept).
\item The system also provides recommendations by analyzing multiple lectures to understand the optimal way to drive the highest engagement in learners.

\end{itemize}

\begin{figure}
    \centering
    \includegraphics[width=\linewidth]{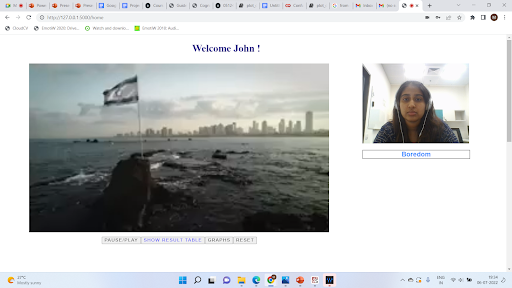}
    \caption{Illustration of Detected Affective State}
    \label{fig:emo_state}
\end{figure}

Tools and technologies used in this work are Python (OpenCV, NumPy, Pandas, Matplotlib, TensorFlow, Keras, sci-kit-learn) and Flask for hosting the Web API. 
Client-side technologies - JavaScript, HTML-5 Canvas, CSS - are used to build a dynamic and user-friendly interface that can capture the user video, and display analytics for the instructor in various formats.

\section{Results}\label{sec6}
In this section, the experimental analysis performed on the proposed approach is discussed. It is done through a comparison of the performance of the proposed method with prior research. This analysis is conducted with the aim of evaluating the proposed system. 

In this section, firstly a comparison of our two proposed models is made- a hybrid CNN-like architecture incorporating focal loss and a hybrid EfficientNet-like architecture incorporating Focal Loss with regard to existing models classifying all four affective states in the dataset. The models employed in this research have fewer trainable parameters and are computationally efficient than VGGs and ResNets. These results are illustrated in Table \ref{table:2}. 

Most of the past works are mainly focused on the classification of only one of the affective states namely, Engagement. The experimental results demonstrate that our proposed method is more effective than other state-of-the-art approaches as it considers the contribution of all affective states as well as provides a comprehensive metric- Attentiveness Index. 
The following results in Table \ref{table:3} contain a comparison of our implemented models with regard to existing methods focusing only on the classification and analysis of the 'Engagement' affective state.

The accuracies obtained by using a CNN-like hybrid model with focal loss are 73.04\%, 79.98\%, 80.17\%, and 85.37\% for boredom, engagement, confusion, and frustration respectively. The highest accuracies obtained by using the EfficientNet-like hybrid model with Focal Loss are 73.16\%, 80.32\%, 79.76\%, and 85.20\% for boredom, engagement, confusion, and frustration respectively.

As shown in Table \ref{table:2}, the proposed method achieves better performance on affective states classification in comparison than existing models.

\begin{table*}
    \caption{Comparison between existing models and proposed methods on classification of all Affective states in DAiSEE}
  \centering
\begin{tabular}{|c|c|c|c|c|}
\hline
\backslashbox{Model}{Affective State} & Boredom & Engagement & Confusion & Frustration \\
\hline
InceptionNet Frame Level \cite{29}
& 36.5\% & 47.1\% & 70.3\% & 78.3\% \\
\hline
InceptionNet Video Level \cite{29}
 & 32.3\% & 46.4\% & 66.3\% & 77.3\% \\
\hline
C3D Training \cite{29}
 & 47.2\% & 48.6\% & 67.9\% & 78.3\%   \\
\hline
C3D FineTuning \cite{29}
 & 45.2\% & 56.1\% & 66.3\% & 79.1\% \\
\hline
LRCN \cite{29}
 & 53.7\% & 57.9\% & 72.3\% & 73.5\%  \\
\hline
EmotionNet \cite{29}
 & 51.07\% & 35.89\% & 57.45\% & 73.09\% \\
\hline
Ours hybrid CNN with FL 
&73.04\% & 79.98\% & 80.17\% & 85.37\% \\
\hline
Ours hybrid EfficientNet with FL
 & 73.16\% & 80.32\% & 79.76\% & 85.20\% \\
\hline

\end{tabular}
\label{table:2}
\end{table*}

\begin{table}
    \centering
    \caption{Comparison of the existing methods with the proposed model for Engagement detection in DAiSEE}
    \renewcommand{\arraystretch}{1.5} % Adjust row height
    \setlength{\tabcolsep}{5pt} % Adjust column spacing
    \begin{tabular}{|p{3cm}|p{3cm}|} % Adjust column width as needed
    \hline
    \textbf{Model} & \textbf{Accuracy} \\
    \hline
    C3D (FL) \cite{33}  & 56.2\% \\
    \hline
    I3D  \cite{34} & 52.5\% \\
    \hline
    DFSTN \cite{35} & 58.84\% \\
    \hline
    DERN \cite{50} & 60.00\% \\
    \hline
    3D DenseAttNet \cite{51} & 63.59\% \\
    \hline
    ViT \cite{52} & 55.18\% \\
    \hline
    Ours hybrid CNN with FL & 79.98\% \\
    \hline
    Ours hybrid EfficientNet with FL & 80.32\%\\
    \hline
    \end{tabular}
\label{table:3}
\end{table}

\section{Conclusion and Future Scope}

Leveraging Artificial Intelligence techniques can enhance e-learning, allowing learners worldwide to study at their own pace and fostering collaborative learning regardless of physical distance or classroom availability. This paper proposes a novel approach for developing a real-time attentiveness prediction system using deep learning and computer vision. In this work, a hybrid architecture using CNN and EfficientNet is developed to classify the affective states among learners in an e-learning environment. This classification method accounts for the contributions of all given affective state classes instead of just focusing on one, unlike most existing works. The proposed method's performance has been evaluated and compared against several prior end-to-end approaches implemented on the DAiSEE dataset, demonstrating optimal improvisation in the results.  A cost-efficient computer vision-based system pipeline is also developed and discussed for learner attentiveness detection which can be integrated with video conferencing platforms or e-learning software, etc. This system enables instructors to receive the aggregated analysis of the entire class to be informed regarding the overall attentiveness pattern of all learners in the context of the learning session.

Future research can examine advanced analysis of head movements, eye movements, and background context as well as information from EEG signals, etc. to enhance the efficiency of gauging attentiveness. To ensure the robustness of this method, a substantial dataset for training and testing purposes that includes a diverse group of learner volunteers, encompassing a wide range of racial backgrounds, while also ensuring data balance can be constructed.

\section*{Declarations}

\begin{itemize}
\item Data availability : \\
The data presented in this study are available on request from the corresponding author. Some data are publicly available and some are not.

\item Conflict of interest/Competing interests : \\
The authors declare that they have no known competing financial interests or personal relationships that could have appeared to influence the work reported in this paper.

\item Funding : \\
The Koret Foundation Grant for Smart Cities and Digital Living 2030.

\item Authors' contributions : \\
Conceptualization, S.G., M.D. P.K. and I.B.G.; methodology, S.G., M.D. P.K. and I.B.G; software, S.G., M.D.; validation, S.G., M.D.; formal analysis, S.G., M.D. P.K. and I.B.G.; investigation, T.R.; resources, P.K. and I.B.G; data curation, S.G., M.D.; writing—original draft preparation, S.G., M.D.; writing—review and editing, P.K. and I.B.G; visualization, , S.G., M.D.; supervision, P.K. and I.B.G; project administration, P.K. and I.B.G; funding acquisition, I.B.G. All authors have read and agreed to the published version of the manuscript.

\item Acknowledgements : \\
This research was partially supported by Koret Foundation Grant for Smart Cities and Digital Living 2030.

\end{itemize}


%\bibliography{citations}

\end{document}